\documentclass[a4paper]{article}

\usepackage{ISCSLP2022}
\usepackage{multirow}

\title{BERT-LID: Leveraging BERT to Improve Spoken Language Identification\thanks{The core source code and data list are available on https://github.com/THUsatlab/BERT-LID.}}
\name{Yuting Nie$^1$$^\dagger$, Junhong Zhao$^2$$^\dagger$ \thanks{$\dagger$ Equal contribution},Wei-Qiang Zhang$^1$$^*$\thanks{* Corresponding author}, Jinfeng Bai$^3$}

\address{$^1$Beijing National Research Center for Information Science and Technology\\
  Department of Electronic Engineering, Tsinghua University, Beijing 100084, China\\
  $^2$Computational Media Innovation Centre, Victoria University of Wellington, New Zealand\\
  $^3$TAL Education, 18 Zhongguancun Avenue, Beijing 100080, China}
\email{nyt19@mails.tsinghua.edu.cn, junhong.jennifer@gmail.com, wqzhang@tsinghua.edu.cn, baijinfeng1@tal.com}
\begin{document}

\maketitle
\begin{abstract}

Language identification is the task of automatically determining the identity of a language conveyed by a spoken
segment. It has a profound impact on the multilingual interoperability of an intelligent speech system. Despite language identification attaining high accuracy on medium or
long utterances($>$3s), the performance on short utterances
($<=$1s) is still far from satisfactory. We propose a BERT-based language identification system (BERT-LID) to improve language identification performance, especially on short-duration speech segments. We extend the original BERT model by taking the phonetic posteriorgrams (PPG) derived from the front-end phone recognizer as input. Then we deployed the optimal deep classifier followed by it for language identification. Our BERT-LID model can improve the baseline accuracy by about 6.5\% on long-segment identification and 19.9\% on short-segment identification, demonstrating our BERT-LID's effectiveness to language identification.

%and ~\highlight{[number]} and F-score
%To get better performance, we use the frame-level Phonetic Posteriorgrams(PPG) derived from the frontend phone recognizer as input, then join different conjunction network after BERT, and finally fine-tune BERT pre-trained the model and the conjunction network.

\end{abstract}

\section{Introduction}

% Automatic speech recognition (ASR) is a language that converts vocabulary in human language into computer cognition. By using better models and making full use of unlabeled data, the ASR mission has made tremendous progress[1][2]. In today's society, everyone generally masters a language, and we hope to process language data that contain multiple languages. Therefore, language identification (LID) can be performed to identify the language type of the input audio, the performance of the system and the effect of the final performance of the ASR on the service[3][4].

% ASR is the conversion of vocabulary in human language into computer-readable language. With the progress and development of society, this is a challenging task. How to quickly and correctly switch between languages is a challenging and meaningful topic[5]. In most ASR systems today, although multilingual speech recognizers can recognize many different spoken languages, it is required that the input test data must contain only one language[6][7]. Therefore, in the absence of language boundaries and language identification information, the performance of the speech recognition task on code switching speech is not as good as the speech recognition task on a single language utterance[8]. Or training a multilingual ASR mod
% el, but this will have higher data requirements[9].

Language identification technology has been widely used to facilitate a variety of speech applications, such as multilingual speech identification~\cite{albadr2019spoken,snyder2018spoken}, spoken language translation~\cite{waibel2008spoken,di2019adapting}, and pronunciation assessment. It enables these speech systems to tackle hybrid-language speech by providing language identities. Considering the case of multilingual speech recognition, a state-of-the-art multilingual speech recognition system is often composed of sub-systems operating in parallel, where each system focuses on a specific language. Language identity can help determine which language model should be triggered to perform the recognition process~\cite{soltau2016neural, battenberg2017exploring}. For the code-switching speech recognition where more than one language is present in one utterance, since there is intra-sentential shifting between language varieties~\cite{lyu2015mandarin}, additional language-specific information (e.g., language switching timestamp) is required by the recognition process to guarantee the accuracy~\cite{nilep2006code,chan2006automatic,li2019towards}. Previous studies have revealed that advances in language identification can contribute to the performance of the speech recognition system~\cite{duroselle2021modeling,liu2021unified}.

\begin{figure}[!b]
\centering % avoid the use of \begin{center}...\end{center} and use \centering instead (more compact)
 \includegraphics[width=0.90\columnwidth]{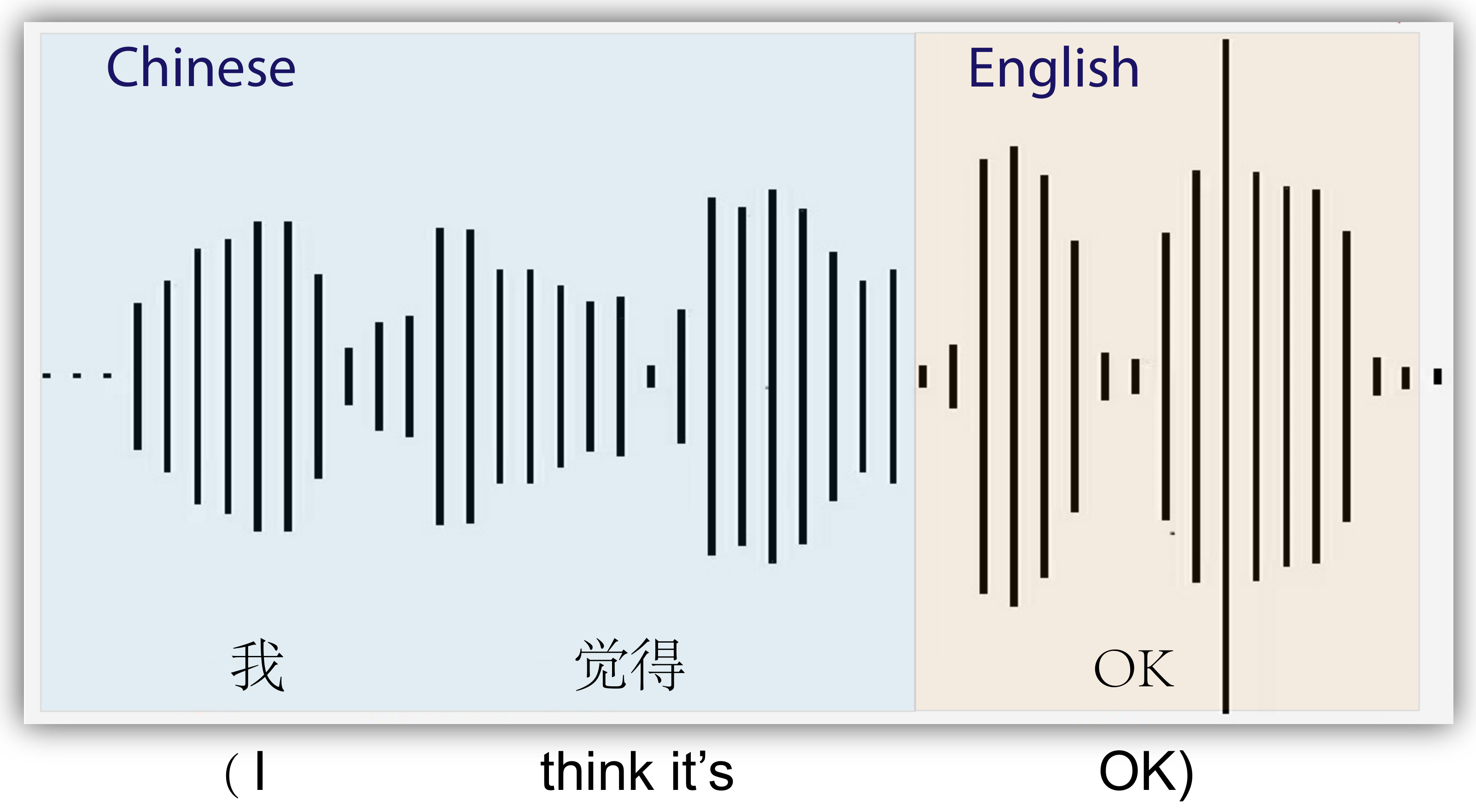}
 \caption{An example of an intra-sentential multilingual utterance, where the Chinese-English language interweaved in one speech segment. The total duration of the utterance is 1.9s, with the first Chinese part last 1s and the second English part last only 0.9s.}
 \label{fig:signal}
\end{figure}

Recent years have witnessed significant improvement in language identifications. Although state-of-the-art solutions achieve high accuracy on medium- and long-length utterances, they remain unsatisfactory for short-length utterances. By short length, we mean less than one second. However, in spontaneous conversation in real life, multilingual code-switching speech occurs quite frequently and often comes up with short-length fragments of a language, as the example shown in Figure~\ref{fig:signal}. The transience characteristic makes it challenging to capture sufficient features and develop effective models. It motivates us to search for more powerful methods to promote the performance of short or even long-segment language identification.  

% Human auditory experiments show that there are two main cues for distinguishing languages: prelexical information and lexical semantic knowledge. Based on this two knowledgements, language identification can be performed. Phonetic repertoire, phonotactics, rhythm, and intonation are all parts of the prelexical information[26].

% Each language contains a phonetic system that controls how symbols are used to form words or morphemes and a syntactic system that controls how words and morphemes are combined to form phrases and utterances. Each language has its own unique phonetic system and syntactic system, which can be used for language identification.

 Languages contain a phonotactic system that dictates how alphabets are used to form words and morphemes, as well as a syntactic system regulating how sentences are constructed with words and morphemes. Every language has its own phonotactic and syntactic systems, which make it distinct from other languages~\cite{yu2003chinese,lyu2008language}. In this paper, we introduce BERT to improve phonotactic and syntactic modeling for language identification, by taking advantage of the enormous capability of BERT on language representation. We investigate how to adapt BERT to model language differentiations. We also explore various BERT combinations with other networks, including CNN, DPCNN, LSTM, and RCNN, for language classification. The performance is evaluated on different datasets in terms of accuracy and F-score. Results indicate that BERT-RCNN is the best-performing model. For both long and short segment language identification, it outperforms the baselines (x-vector~\cite{snyder2018x} and n-gram-SVM~\cite{wang2006multi}) by a large margin on accuracy and F1-score, with the most improvement on the short-segment language identification task.

%and \highlight{[TBD\%]}5$\%$ on F-score

% In our work, we mainly studied how to apply the BERT model to the field of speech. This is the first time that the BERT model has been applied to the field of LID. In the model, try different networks such as CNN, DPCNN, LSTM, RCC, etc. to compare the results. The purpose of our work is to solve the problem of improving the accuracy of language identification in often short audio. Through our experiments, we found that the language identification system based on the BERT model can achieve higher accuracy, although the performance will be slightly different with the different back-end network structures.

\section{Related Work}
\subsection{Language identification}
The language identification problem has been extensively investigated in the speech domain. There are acoustic model methods and phonotactic methods that constitute mainstream techniques~\cite{snyder2018spoken,song2015deep,zhan2021self}. The acoustic model methods concentrate on modeling acoustic spectral features. The phonotactic method, however, often starts by using single/multiple phone recognizer to decode speech sequences into one/multiple phone sequences, then goes on to model phono- or syntactic relationships from there. For acoustic model approaches, the i-vector based system has led the way for a long time~\cite{dehak2010front}, and then the x-vector based system followed~\cite{snyder2018x}. Both systems have two main components, acoustic feature embedding extractor, and back-end classifier. X-vector innovates i-vector system with a discriminative feature embedding extractor including deep neural network and its counterparts~\cite{snyder2018spoken,song2015deep}. For phonotactic approaches, major efforts have been taken on language model improvement, beginning with the seminal work of the n-gram lexicon model, to vector space model (VSM)~\cite{li2006vector,zhang2014spoken,lee2020subspace} and recent work on RNN~\cite{salamea2016use, salamea2018language} and transformer-based language models~\cite{romero2021exploring}. BERT-LID will follow the phonotactic research track. Unlike prior works, it will incorporate not only phonotactic but also syntactic-level language discriminations to improve long- and particularly short- segment language identification.

%~\highlight{TODO: highlight some representative work in this directions}.

%For phonotactic-based approaches, ~\highlight{TODO: highlight some representative work in this directions}. BERT-LID will follow the phonotactic-based research track and will compare with both x-vector and PRLM system.

\subsection{BERT}
BERT is a powerful transformer-based encoder pre-trained using masked language modeling and the next sentence predicting objectives to effectively model long contextual dependencies of the input text~\cite{devlin2018bert}. Originally, BERT was meant to work with text-based representations. However, we will swap out the BERT's input to suit speech-based applications.
%However, we will work on the first layer of BERT to suit speech-based applications.

\subsection{BERT in language identification}
% ~\highlight{BERT-related work on language identification, add paper mentioned}
The prior work~\cite{zhan2021self} applied the BERT-style pertaining  concept to improve language identification. However, it mainly focus on acoustic-based methods, to partially improve the acoustic features extractions of the x-vector system by training with time and frequency-wisely masking objectives. In contrast, BERT-LID positions the BERT as a phonotactic-syntactic feature extractor (language model), capable of modeling phoneme and higher level information.

\section{System Description}

% Previous studies have shown that the BERT model has achieved cross-century achievements in NLP tasks. For speech and text, there is a great similarity, so this is done on the basis of BERT to apply the BERT model to speech tasks.
\begin{figure*}[!ht]
\centering % avoid the use of \begin{center}...\end{center} and use \centering instead (more compact)
 \includegraphics[width=1.9\columnwidth]{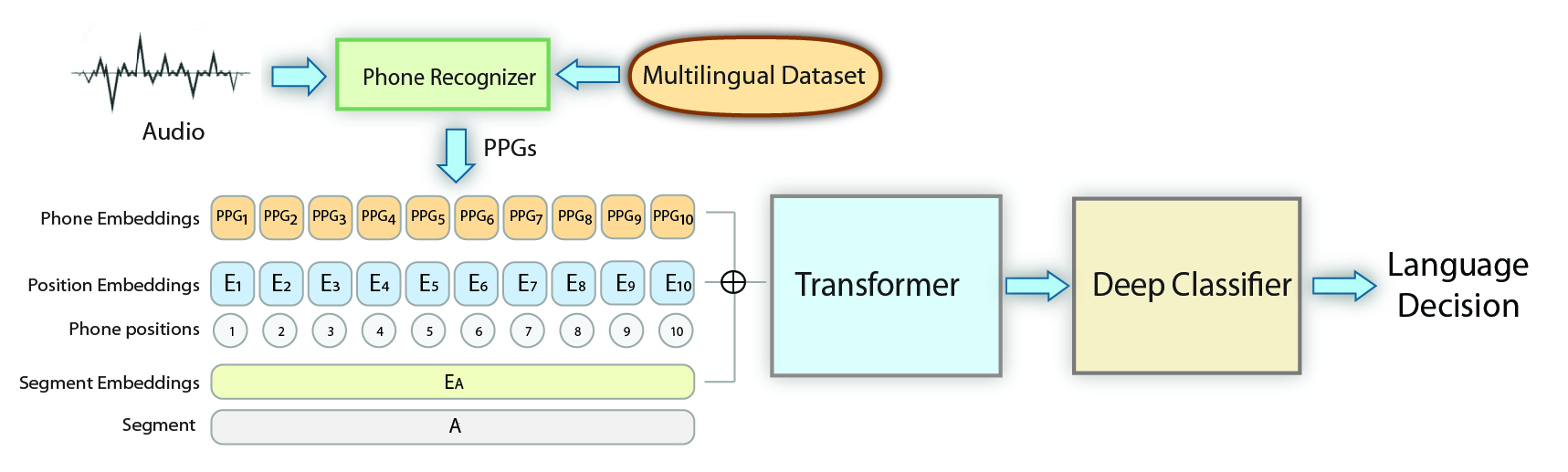}
 \caption{The scheme of BERT-LID. It features a customized BERT module to extract language-specific phonotactic representations and a deep classifier for language identification. In particular, we used the phone-wise averaged PPGs extracted from the phone recognizer as the token embeddings of BERT.
 We used RCNN~\cite{lai2015recurrent} as the deep classifier, following BERT.
 }
 \label{fig:scheme}
\end{figure*}

The proposed work aims to improve the language identification performance on both long and short-length speech segments, mainly focusing on exploring advanced phonotactic and syntactic modeling methods. As a means of achieving this goal, we take advantage of BERT's capabilities for language representation and explore its benefit for language identification. Our proposed BERT-LID system is illustrated in Figure~\ref{fig:scheme}. It includes a phone recognizer, a customized BERT module, and a deep classifier.

\subsection{Phone recognizer}
 Following the general phonotactic-based language identification approaches, our pipeline begins with a phone recognizer, developed by using the bottleneck NN network trained with block softmax output layer~\cite{fer2017multilingually}. It takes Fbank features and outputs the phonetic recognition results. Rather than directly using recognized phonetic transcriptions for the following language modeling, we rely on the Phonetic Posterior Grams (PPG), which has been widely used in a broad spectrum of application~\cite{hazen2009query,sivaram2011multilayer,fer2017multilingually}. For each frame of the running speech, it gives the posterior probability belonging to each phone or phone-like unit. The phone recognizer will remain fixed once it has been built.
%[~\highlight{[TBD][methods/network structure used in phone recognizer, and citations]}]

% Backend classifiers identify languages based on the language features derived from BERT. We investigate several variants of the classifier by taking into account representative networks from the sequence processing domain (see Sec.~\ref{sec:classifier}).

\subsection{BERT adaption and token embedding methods}
\label{sec:token embedding}

Following the phone recognizer is the customized BERT module. The inputs of the original BERT model are the aggregation of the token (word/subword), position, and segment embeddings. In this work, we study how to adapt BERT to the phone-level task at a low cost. We chose to maintain the same triplet aggregation scheme as the original BERT but change primarily the token and position embeddings from token-level to frame- and phone-level, depending on the front-end phone recognition output. We explore different embedding methods, including:

(1) Frame-level PPGs and position embeddings (denoted as \textbf{PPG$_{frm}$} ), in which we use the PPG vectors of each frame as the frame embedding vectors. Position embeddings are derived from frame positions spread throughout the whole segment.

(2) Text-based phone and position embeddings (denoted as \textbf{Phone$_{emb}$}), in which we obtain first the phone recognition results from the front-end phone recognizer, and then use the transcription to extract the phone embeddings based on a lookup table, as in the original BERT algorithm~\cite{devlin2018bert}. The phone positions arranged within the whole segment are used to extract the position embeddings.

(3) The phone-wise averaged PPGs and position embeddings (denoted as \textbf{AvPPG} ), where the phone embeddings are obtained by averaging the PPGs based on the time boundary extracted from the phone recognizer. As (2), the position embedding is derived based on the phone positions along with the entire segment.

% in Table~\ref{tab:token_embedding_OLR20} and~\ref{tab:token_embedding_Timit_THCHS30}
%in Table~\ref{tab:token_embedding_OLR20} and~\ref{tab:token_embedding_Timit_THCHS30}
% in Table~\ref{tab:token_embedding_OLR20} and~\ref{tab:token_embedding_Timit_THCHS30}
% \label{sec:token embedding}
% We explore different token embeddings fed into the BERT module to see if the token form matters to the final performance. Three token forms are considered:
% (1) Frame-level PPGs and positions (denoted as \textbf{PPG$_{frm}$} in Table~\ref{tab:token_embedding_OLR20} and~\ref{tab:token_embedding_Timit_THCHS30}), in which we use the PPG vectors of each frame as the token embeddings, together with the incremental frame positions spread within the whole segment as the BERT input.
% (2) Text-based phone embedding and positions(denoted as \textbf{Phone$_{emb}$} in Table~\ref{tab:token_embedding_OLR20} and~\ref{tab:token_embedding_Timit_THCHS30}), in which we obtain phone recognition results directly from the front-end phone recognizer, then used it to extract the text-based phone embeddings, together with the phone positions arranged within the whole segment as the input.
% (3) The phone-wise averaged PPGs and positions (denoted as \textbf{AvPPG} in Table~\ref{tab:token_embedding_OLR20} and~\ref{tab:token_embedding_Timit_THCHS30}), where the token embeddings are obtained by averaging the PPGs based on the time boundary extracted from the phone recognizer. As (2), the phone positions alongside the whole segments are combined as input.

\subsection{Deep classifier variants}
\label{sec:classifier}
% \begin{itemize}%[leftmargin=*]
% \item \textbf{BERT-CNN}
% \end{itemize}
We investigate several variants of the classifier that follow the BERT module for language identification. The representative networks from the sequence processing domain are taken into account, including:

% \textbf{CNN: }
% \noindent{\textbf{CNN}~\cite{krizhevsky2012imagenet} is a widely-used network structure in both speech and image processing domains. It has outstanding feature modeling capabilities by considering both spatial and temporary information. The CNN classifier we used include one 1-D convolutional layer with the kernel size of 200, followed by a max-pooling and a linear layer.}

\noindent{\textbf{CNN}~\cite{krizhevsky2012imagenet}. The CNN classifier we used include one 1-D convolutional layer with the kernel size of 200, followed by a max-pooling and a linear layer.}

%PPG in phoneme-order ~\highlight{[TBD][How to feed input]} and feed it to the

% \begin{itemize}%[leftmargin=*]
% \item \textbf{BERT-LSTM}
% \end{itemize}
% \noindent{\textbf{LSTM}~\cite{hochreiter1997long} stands out when it comes to sequential modeling. Compared with other sequential modeling methods, LSTM reasonably addresses the gradient disappearance and explosion issues during long-sequence training and thus can better model long-range dependencies. The LSTM classifier we utilized include two bidirectional hidden layers with 300 hidden nodes, followed by a max-pooling and a linear layer.  
% }

\noindent{\textbf{LSTM}~\cite{hochreiter1997long}. The LSTM classifier we utilized include two bidirectional hidden layers with 300 hidden nodes, followed by a max-pooling and a linear layer.  
}

%~\highlight{[TBD][How to feed input, sentence level? Or frame-level chuncks?]}.

% \begin{itemize}%[leftmargin=*]
% \item \textbf{BERT-DPCNN}
% \end{itemize}
\noindent{\textbf{DPCNN} (Deep Pyramid Convolutional Neural Networks)~\cite{johnson2017deep}. Following Johnson et al.~\cite{johnson2017deep}, the DPCNN network we use has a region embedding extractor followed by repeating convolution blocks, each of which includes one 1/2 pooling layer and two equal-width convolution layers coupled with skip connections (15 weight layers in all).}

% \begin{itemize}%[leftmargin=*]
% \item \textbf{BERT-RCNN}
% \end{itemize}
% \noindent{\textbf{RCNN} (Recurrent Convolutional Neural Networks)~\cite{lai2015recurrent} can capture long-range word-level contextual information and has been successfully used in text classification. Unlike the original RCNN, here we use it to classify speech languages based on BERT features. The RCNN we use contains a bidirectional LSTM followed by a max-pooling layer. From the bidirectional LSTM, left and right context vectors are concatenated to represent the current frame in its context.}

\noindent{\textbf{RCNN} (Recurrent Convolutional Neural Networks)~\cite{lai2015recurrent}. The RCNN we use contains a bidirectional LSTM followed by a max-pooling layer. From the bidirectional LSTM, left and right context vectors are concatenated to represent the current frame in its context.}

A softmax layer was added to each classifier as the penultimate layer for classification. The classifier is initialized with random initialization method and is trained with a cross-entropy loss.

% During our experiment, we first changed BERT, which mainly manifested itself in changing its input from text form to speech tokens. On the BERT-base-cased model provided by BERT, we used our data for fine-tuning to complete the language identification process.

% In the NLP task, the input is all text, which is tokenized by BERT and then processed. For speech tasks, the input is tokens, so there is no need to add an additional output layer at the same time as the initial text change token. To make fine-tuning to complete the text-to-speech migration.

% In the BERT model, the representation in the layer can be trained by jointly adjusting all channels. Therefore, the preset BERT means that it is suitable for fine-tuning through an additional output layer and is suitable for the construction of the most advanced model for a wide range of tasks. To complete the previous LID tasks that were not systematically studied or presented in the BERT model, we added different neural networks for fine-tuning. 

% We consider the current relatively successful network structure and the network structure suitable for sequence processing. To better solve the sequence problem, we should use the timing relationship. Therefore, we tend to use RNN-based networks and their variants to train LER tasks and compare their performance in experiments. The detailed comparison results will be described in the next chapter.
\section{Experiments}
We evaluate BERT-LID on long- and short-segment language identification tasks, including natural short-segment code-switching speech. Considering the binary classification tasks on such code-switching data, we use not only traditional EER but also accuracy and F1-score as evaluation metrics. 

\subsection{Baseline systems}
We compared the BERT-LID system with one phonotactic-based baseline n-gram-SVM and the state-of-the-art acoustic-based baseline, x-vector. The phonotactic-based baseline starts with a phone recognizer, followed by a bag-of-n-grams model to capture the phonotactic features and an SVM as a classifier. For acoustic-based method comparison, we follow the prior work~\cite{li2020ap20} to set up the x-vector baseline system. 

% The input features are 23-dimensional MFCCs extracted under a frame length of 25ms.

% We used 3-gram statistics based on the phoneme-level occurrence frequency, the TF-IDFs~\cite{zhang2010time}, and phoneme duration as features.

\subsection{Datasets}
We use three datasets in our experiments for training and evaluation. OLR20 is a multilingual dataset containing six languages. The average segment length of this dataset is 5.45s. {T\&T} is a Chinese-English mixture dataset combining THCHS-30 (Chinese)~\cite{wang2015thchs} and TIMIT (English)~\cite{garofolo1993darpa}. We made it a short-length dataset by chopping all sentences (non-silent part) into one-second speech fragments. TAL$\_$ASR is a dataset collected under a scenario of English classes taught by Chinese. It contains many code-switching speeches with Chinese and English intertwined. 

% To obtain enough short segments for testing, we first conduct the force alignment and segment the long utterance into pieces by the ground-truth languages labels. Then we chose the segments within 1s long to be our testing data. 100 thousands pieces of samples with a ratio of 1:1 between Chinese and English are selected in the end as the testing data.

% Each of the three datasets is divided into training and testing by a 1:9 ratio in our experiments.
%~\highlight{[TBD]How to chop? Have you considered semantic endpoint. Or just randomly chop (if so, a lot of broken word will appear )? Overlap in the segments? }.
\vspace{-0.4cm}
\begin{table}[th]
  \caption{Dataset overview.}
  \label{tab:example}
  \centering
%   \vspace{0.2cm}
  \scalebox{0.95}{
  \begin{tabular}{ccccc}
    \toprule
    \multicolumn{1}{c}{\textbf{Dataset}} & \multicolumn{1}{c}{\textbf{Length}}&
    \multicolumn{1}{c}{\textbf{Languages}}\\
    \midrule
OLR20  &  Long & Multilanguage \\
T\&T  &  Short (1s) & Bilanguage \\
TAL\_ASR &  Short (Code-switching) & Bilanguage \\
    \bottomrule
  \end{tabular}
  }
\label{tab:Dataset}

\end{table}
 \vspace{-0.4cm}

\subsection{Implementation detail}
%~\highlight{[TBD] trainng epoches, gradient}

% Our implementation is based on PyTorch~\cite{paszke2019pytorch}. 

To train models, we use the BertAdam optimizer with learning rate of $5 \times 10^{-5}$, $\beta_1=0.9$,  $\beta_2=0.99$. The training epochs range between [150, 400], and gradient accumulation steps range between [1,4]. They will be chosen case by case based on the performance on the development dataset. We will forcefully terminate the training when the loss in the development dataset successively increases by 50 times. For more implementation details of our experiments, we encourage readers to refer to our code.

\section{Results}
\subsection{Phone recognition impact}
We use the open-source tool BUT~\cite{fer2017multilingually} to develop the phone recognizer and extract PPGs. It provides numerous models trained on different datasets, including monolingual Fisher-English model and multilingual Babel-17 (17 languages) model. We experiment with both models on BERT-CNN and BERT-RCNN pipelines to validate the contributions of the phone recognizer for the final performance.

The results are shown in Table~\ref{tab:phone_recongition}. It's observed that different phone recognizers will impact the final performance on both accuracy and F1-score, although the significance varies across different system setups. The Babel-17 model showed adequate robustness. As such, we will use it for PPG extractions in the following experiments.
%~\highlight{Insights of the numbers and take-aways}.

\begin{table}[th]
  \caption{Performance of BERT-CNN and BERT-RCNN with different phone recognizers on OLR20 datset.}
  \label{tab:phone_recongition}
  \centering
    \resizebox{0.45\textwidth}{!}{ 
  \begin{tabular}{ccccc}
    \toprule
    \centering
    {\textbf{Model}}&
    {\textbf{Training dataset}} &
    {\textbf{Accuracy}}&
    {\textbf{F1-score}}\\
    
    \midrule
    
    \multirow{2}{*}{BERT-CNN} & FisherEnglish & 0.9760 & 0.9760 \\ ~  & Babel-17&  0.9716 & 0.9724 \\
    \midrule 
    \multirow{2}*{BERT-RCNN} & FisherEnglish & 0.9483 & 0.9479 \\ ~ & Babel-17&  {\bf 0.9763} & \textbf{0.9764} \\
    \bottomrule
  \end{tabular}}
\end{table}
% \begin{table}[th]
%   \caption{A comparison of BERT-CNN and BERT-RCNN systems on the OLR20 dataset using different phone recognition model. }
%   \label{tab:phone_recongition}
%   \centering
%     \scalebox{0.95}{
%   \begin{tabular}{ccccc}
%     \toprule
%     \centering
%     {\textbf{Model}}&
%     {\textbf{Phone Recognizer}} &
%     {\textbf{Accuracy}}&
%     {\textbf{F1-score}}\\
%     \midrule
%     \multirow{2}{*}{BERT-CNN} & FisherEnglish  & 0.9798 & 0.9798 \\ ~  & Babel-17  &  0.9792 & 0.9792 \\
%     \multirow{2}*{BERT-RCNN} & FisherEnglish  & 0.9618 & 0.9620 \\ ~ & Babel-17&  {\bf 0.9871} & \textbf{0.9873} \\
%     \bottomrule
%   \end{tabular}
%   }
% \end{table}

\subsection{Token embedding impact}

\begin{table}[th]
  \caption{Accuracy of different token embedding methods for BERT module on OLR20 and T\&T dataset}
  \centering
    \resizebox{0.45\textwidth}{!}{ 
  \begin{tabular}{*{5}{c}}
    \toprule
    {\textbf{Database}}&
    {\textbf{Model}} &
    {\textbf{PPG$_{frm}$}}&
    {\textbf{Phone$_{emb}$}}&
    {\textbf{AvPPG}}
    \\

    %\multicolumn{1}{c}{\textbf{Database} & 
    % \multirow{2}{*}{Model(Acc)}&
    % \multicolumn{2}{c}{\textbf{PPG$_{frm}$}}&
    % \multicolumn{2}{c}{\textbf{Phone$_{emb}$}}&
    % \multicolumn{2}{c}{\textbf{AvPPG}}\\

    % \cmidrule{2-7} & OLR20 & T1 & OLR20 & T1 & OLR20 & T1\\

    % \midrule
    % CNN  & 0.9792 & 0.9792 & 0.9792 & 0.9792 & 0.9792 & 0.9792 \\
    
    % \multicolumn{1}{c}{\textbf{Model(Acc)}} & 
    % \multicolumn{1}{c}{\textbf{PPG$_{frm}$}}&
    % \multicolumn{1}{c}{\textbf{Phone$_{emb}$}}&
    % \multicolumn{1}{c}{\textbf{AvPPG}}\\
    \midrule

    \multirow{4}{*}{OLR20} & CNN  &  0.9792& 0.9734 &\textbf{0.9821}\\~ & LSTM  &  0.9204 & 0.9308  & \textbf{0.9369}\\~ & DPCNN &  0.9634 & 0.9669 & \textbf{0.9736}\\~ & RCNN &  0.9871 & 0.9835 &\textbf{0.9921} \\

    \midrule 

    \multirow{4}*{T\&T} & CNN  &  0.9198& 0.9272 & \textbf{0.9306}\\
~ & LSTM  &  0.9327 & 0.9408  & \textbf{0.9484}\\
~ & DPCNN &  0.9302 & 0.9482 & \textbf{0.9524}\\
~ & RCNN &  0.9623 & \textbf{0.9771 }&0.9751 \\

    \bottomrule
  \end{tabular}}
\label{tab:token_embedding_OLR20}
\end{table}

% \begin{table}[th]
%   \caption{Evaluation of different token embeddings input into BERT module on TIMIT\&THCHS30 (1s) dataset.}
%   \centering
%   \begin{tabular}{ccccc}
%     \toprule
%     \multicolumn{1}{c}{\textbf{Model(Acc)}} &
%     \multicolumn{1}{c}{\textbf{PPG$_{frm}$}}&
%     \multicolumn{1}{c}{\textbf{Phone$_{emb}$}}&
%     \multicolumn{1}{c}{\textbf{AvPPG}}\\
%     \midrule
% CNN  &  0.9198& 0.9272 & \textbf{0.9306}\\
% LSTM  &  0.9327 & 0.9408  & \textbf{0.9484}\\
% DPCNN &  0.9302 & 0.9482 & \textbf{0.9524}\\
% RCNN &  0.9623 & \textbf{0.9771 }&0.9751 \\
%     \bottomrule
%   \end{tabular}
% \label{tab:token_embedding_Timit_THCHS30}
% \end{table}
 
We evaluate the performance of different token embedding methods on the final language identification results. As shown in Table~\ref{tab:token_embedding_OLR20} 
%and Table ~\ref{tab:token_embedding_Timit_THCHS30}
, the network performs better when the token embedding changed from frame-level \textbf{(PPG$_{frm}$}, 2nd column) to phone-level (\textbf{Phone$_{emb}$} and \textbf{AvPPG}, 3rd and 4th column). The phone-wise averaged PPGs achieve the best performance in most cases. One possible underlying reason is that phone-level input has a reduced mismatch with the BERT pre-trained model that trained on word- or sub-word language tasks compared to frame-level input. It's noteworthy that for \textbf{Phone$_{emb}$} case, it needs to keep the phoneme set used by the phone recognizer to be consistent with the language dictionary used by BERT pre-training to obtain the text-based phone embedding.

%The generated arbitrary representation will likely be less than the input at the frame-level. 
\subsection{Ablation study}

We combined BERT modules with the deep classifier to improve language identification. To understand the contribution of each module, we show how our network performs with either just a BERT (denoted as \textbf{BERT}) or a deep classifier (denoted as \textbf{LID}). In \textbf{BERT} case, one output layer was added to the BERT pre-trained model and fine-tuned for language classification. In \textbf{LID} case, the PPGs are directly fed into the deep classifier. With the frame-level PPGs as the input, we evaluated different networks on the OLR20 and {T\&T} datasets. The accuracy is shown in Table ~\ref{tab:Ablation_all_data} 
%and Table ~\ref{tab:Ablation_TIMIT_THCHS30}
.
\begin{table}[th]
  \caption{Accuracy of different models in ablation studies on OLR20 and T\&T dataset.}
  \centering
  \resizebox{0.45\textwidth}{!}{ 	
  \begin{tabular}{*{5}{c}}
    \toprule
    {\textbf{Database}}&
    {\textbf{Model}} &
    {\textbf{BERT-LID}}&
    {\textbf{LID}}&
    {\textbf{BERT}}
    \\
    \midrule
    \multirow{4}{*}{OLR20} & CNN  &  \textbf{0.9792} & 0.9412&\multirow{4}{*}{0.9183} \\
&LSTM  &  \textbf{0.9204} & 0.9318 \\
&DPCNN &  \textbf{0.9634} &  0.9572\\
&RCNN & \textbf{ 0.9871} & 0.9619 \\

    \midrule 
    
    \multirow{4}*{T\&T} & CNN  &  \textbf{0.9198} & 0.8295 & \multirow{4}{*}{0.8096}\\
&LSTM  & \textbf{ 0.9327} & 0.8705 \\
&DPCNN &  \textbf{0.9302} &  0.8616\\
&RCNN &  \textbf{0.9623} & 0.8738 \\

    \bottomrule
  \end{tabular}}
\label{tab:Ablation_all_data}
\end{table}

We found that using BERT representations can consistently improve performance compared with directly feeding frame-level PPGs into the back-end classifier. On OLR20, BERT-RCNN increases the identification accuracy from 96.19\% to 98.71\%. More improvements were achieved on short-length data of {T\&T}. Results demonstrate that BERT is compatible with phone- or frame-level PPGs and can generate the language representation reliably for language identification.

% \begin{table}[th]
%   \caption{Ablation study on OLR20 dataset.}
%   \centering
%   \begin{tabular}{ccccc}
%     \toprule
%     \multicolumn{1}{c}{\textbf{Model(Acc)}} & \multicolumn{1}{c}{\textbf{BERT-LID}}&
%     \multicolumn{1}{c}{\textbf{LID}}&
%     \multicolumn{1}{c}{\textbf{BERT}}\\
%     \midrule
% CNN  &  \textbf{0.9792} & 0.9412&\multirow{4}{*}{0.9183} \\
% LSTM  &  \textbf{0.9204} & 0.9318 \\
% DPCNN &  \textbf{0.9634} &  0.9572\\
% RCNN & \textbf{ 0.9871} & 0.9619 \\
%     \bottomrule
%   \end{tabular}
% \label{tab:Ablation_OLR20}
% \end{table}

% \begin{table}[th]
%   \caption{Ablation study on TIMIT\&THCHS30 (1s) dataset.}
%   \centering
%   \begin{tabular}{ccccc}
%     \toprule
%     \multicolumn{1}{c}{\textbf{Model(Acc)}} & \multicolumn{1}{c}{\textbf{BERT-LID}}&
%     \multicolumn{1}{c}{\textbf{LID}}&
%     \multicolumn{1}{c}{\textbf{BERT}}\\
%     \midrule
% CNN  &  \textbf{0.9198} & 0.8295 & \multirow{4}{*}{0.8096}\\
% LSTM  & \textbf{ 0.9327} & 0.8705 \\
% DPCNN &  \textbf{0.9302} &  0.8616\\
% RCNN &  \textbf{0.9623} & 0.8738 \\
%     \bottomrule
%   \end{tabular}
% \label{tab:Ablation_TIMIT_THCHS30}
% \end{table}

\begin{table}[th]
  \caption{Comparision results on OLR20, T\&T, and TAL$\_$ASR (code-switching short) dataset.}
  \centering
    \resizebox{0.45\textwidth}{!}{ 
  \begin{tabular}{*{5}{c}}
    \toprule
    {\textbf{Database}}&
    {\textbf{Model}} &
    {\textbf{EER}}&
    {\textbf{Accuracy}}&
    {\textbf{F1-score}}
    \\
    \midrule
    \multirow{3}{*}{OLR20} & BERT-RCNN &\textbf{0.0218} &  \textbf{0.9921} & \textbf{0.9951} \\
~ & n-gram-SVM &0.0976 &  0.9268 & 0.9260 \\
~ & x-vector& 0.1670&  0.8590 & 0.8592\\

    \midrule 
    
    \multirow{3}*{T\&T} & BERT-RCNN  &\textbf{0.0582}&  {\bf 0.9751} & \textbf{0.9746} \\
~ & n-gram-SVM &0.1736 &  0.8572 & 0.8572 \\
~ & x-vector &0.3506 &  0.6971 & 0.6971\\

    \midrule 
    
    \multirow{2}*{TAL\_ASR} & BERT-RCNN & {\bf0.0813} &  {\bf 0.9461} & \textbf{0.9456} \\
~ & n-gram-SVM& 0.2935 &  0.7468 & 0.7460\\

    \bottomrule
  \end{tabular}}
\label{tab:all_comparision}
\end{table}

% \begin{table*}[h]
%   \caption{Evaluation of different token embeddings input into BERT module on datasets.}
%   \centering
%   \begin{tabular}{*{10}{c}}
%     \toprule
 
%     \multirow{2}{*}{Model}&
%     \multicolumn{3}{c}{\textbf{OLR20}}&
%     \multicolumn{3}{c}{\textbf{TT1s}}&
%     \multicolumn{3}{c}{\textbf{TALASR}}\\
%     \cmidrule{2-10} & \textbf{EER} & \textbf{Accuracy} & \textbf{F1-score} & \textbf{EER} & \textbf{Accuracy} & \textbf{F1-score} & \textbf{EER} & \textbf{Accuracy} & \textbf{F1-score} \\

%     \midrule
%     BERT-RCNN &\textbf{0.0218} &  \textbf{0.9921} & \textbf{0.9951} & \textbf{0.0582}&  {\bf 0.9751} & \textbf{0.9746} & TODO &  {\bf 0.9461} & \textbf{0.9456} \\

%     n-gram-SVM baseline&0.0976 &  0.9268 & 0.9260 & 0.1736 &  0.8572 & 0.8572 & TODO &  0.7468 & 0.7460 \\

%     x-vector baseline& 0.1670&  0.8590 & 0.8592 & 0.3506 &  0.6971 & 0.6971 & TODO &  TODO & TODO \\

%     \bottomrule
%   \end{tabular}
% \label{tab:all_comparision}
% \end{table*}

\subsection{Comparisions}
% \label{sec:eval_OLR20}
% \begin{itemize}[leftmargin=*]
% \item \textbf{OLR20}
% \end{itemize}
From the above experiments, we found that the BERT-RCNN model with phone-wise averaged PPGs shows the best performance in both long and short-duration conditions. So in our comparisons, we choose this system and compare it with the baseline systems.

\noindent{\textbf{OLR20.}}
From Table~\ref{tab:all_comparision} we can see that BERT-RCNN systems outperform both baseline systems by a large margin on OLR20 dataset. It yields a 0.08 EER, 6.5\% accuracy and 6.9\% F-score improvements compared with n-gram-SVM baseline, and 0.15 EER, 13\% accuracy and F-score improvement compared with the x-vector baseline. The x-vector system shows no advantage over other systems, obtaining the lowest accuracy and F-score values.
% we can see that the greatest performance
% gain is often obtained in the first iteration.

% \begin{table}[th]
%   \caption{~\highlight{ORIGINAL}Results of different BERT-LID system on OLR20 datset.}
%   \label{tab:result_olr20}
%   \centering
%   \begin{tabular}{ccccc}
%     \toprule
%     \multicolumn{1}{c}{\textbf{Model}} & \multicolumn{1}{c}{\textbf{Accuracy}}&
%     \multicolumn{1}{c}{\textbf{F1-score}}\\
%     \midrule
% BERT-CNN   &  0.9716 & 0.9724 \\
% BERT-LSTM  &  0.8949 & 0.8954 \\
% BERT-DPCNN &  0.9595 & 0.9595 \\
% BERT-RCNN  &  \textbf{0.9763} & \textbf{0.9764} \\
% n-gram-SVM baseline &  0.9268 & 0.9260 \\
% x-vector baseline &  0.8590 & 0.8592\\
%     \bottomrule
%   \end{tabular}
% \end{table}

% \begin{table}[t!]
%   \caption{Comparison results on OLR20 datset.}
%   \label{tab:result_olr20}
%   \centering
%   \begin{tabular}{ccccc}
%     \toprule
%     \multicolumn{1}{c}{\textbf{Model}} &
%     \multicolumn{1}{c}{\textbf{EER}}&
%     \multicolumn{1}{c}{\textbf{Accuracy}}&
%     \multicolumn{1}{c}{\textbf{F1-score}}\\
%     \midrule
% % BERT-CNN   &  0.9792 & 0.9792 \\
% % BERT-LSTM  &  0.9204 & 0.9205 \\
% % BERT-DPCNN &  0.9634 & 0.9635 \\
% BERT-RCNN &\textbf{0.0218} &  \textbf{0.9921} & \textbf{0.9951} \\
% n-gram-SVM baseline&0.0976 &  0.9268 & 0.9260 \\
% x-vector baseline& 0.1670&  0.8590 & 0.8592\\
%     \bottomrule
%   \end{tabular}
% \end{table}

\noindent{\textbf{T\&T.}}
% \subsection{Evaluations on TIMIT$\&$THCHS30}
The evaluation on T$\&$T dataset is to explore the BERT-LID performance on short-segment language identification tasks. As we can observe, there is a significant performance degradation across all systems on this dataset compared with OLR20 results. Among them, the x-vector system has the largest drops. Such a result is predictable since the one-second constraints are rather restrictive for sufficient feature extraction and decision-making. Although this, the BERT-LID model can still improve the baseline systems significantly, giving a 11.8\% improvement on both metrics (from 85.72\% on n-gram-SVM to 97.51\% on BERT-RCNN). It reflects our assumption that improving phonological and syntactic representations that BERT-LID mainly contributes to will benefit short-segment language identification.

% \begin{table}[th]
%   \caption{~\highlight{ORIGINAL}Results of different BERT-LID system on TIMIT\&THCHS30 dataset.}
%   \label{tab:result_timit_thchs30}
%   \centering
%   \begin{tabular}{ccccc}
%     \toprule
%     \multicolumn{1}{c}{\textbf{Model}} & \multicolumn{1}{c}{\textbf{Accuracy}}&
%     \multicolumn{1}{c}{\textbf{F1-score}}\\
%     \midrule
% BERT-CNN   &  0.9035 & 0.9036 \\
% BERT-LSTM  &  0.9140 & 0.9114 \\
% BERT-DPCNN &  0.9160 & 0.9164 \\
% BERT-RCNN  &  {\bf 0.9460} & \textbf{0.9461} \\
% n-gram-SVM baseline &  0.8572 & 0.8572 \\
% x-vector baseline &  0.6971 & 0.6971\\
%     \bottomrule
%   \end{tabular}
% \end{table}

% \begin{table}[t!]
%   \caption{Comparision results on TIMIT\&THCHS30 (1s) dataset.}
%   \label{tab:result_timit_thchs30}
%   \centering
%   \begin{tabular}{ccccc}
%     \toprule
%     \multicolumn{1}{c}{\textbf{Model}} & \multicolumn{1}{c}{\textbf{EER}}&
%     \multicolumn{1}{c}{\textbf{Accuracy}}&
%     \multicolumn{1}{c}{\textbf{F1-score}}\\
%     \midrule
% % BERT-CNN   &  0.9198 & 0.9198 \\
% % BERT-LSTM  &  0.9327 & 0.9330 \\
% % BERT-DPCNN &  0.9302 & 0.9302 \\
% BERT-RCNN  &\textbf{0.0582}&  {\bf 0.9751} & \textbf{0.9746} \\
% n-gram-SVM baseline&0.1736 &  0.8572 & 0.8572 \\
% x-vector baseline&0.3506 &  0.6971 & 0.6971\\
%     \bottomrule
%   \end{tabular}
% \end{table}

\noindent{\textbf{TAL$\_$ASR.}}
% \subsection{Evaluations on TAL$\_$ASR}
We also evaluate the BERT-RCNN system on the bilingual code-switching dataset TAL$\_$ASR. As shown in Table~\ref{tab:all_comparision}, the intra-sentential multilingual recognition task present in TAL$\_$ASR is the most challenging task and got the lowest performance compared with OLR20 and T\&T. However, the BERT-RCNN system showed the most significant improvement over the n-gram-SVM baseline, by a margin of 19.9\% on both accuracy and F1-score, and 0.21 on EER. Those findings indicate that BERT-RCNN has the most excellent generalizability and can highly benefit language identification of short-segment speech from real-life scenarios.
%\highlight{How many of such data? The percentage of each language? Should we test on the original dataset as well?}.

% \begin{table}[th]
%   \caption{~\highlight{ORIGINAL}Comparisions of BERT-RCNN with n-gram-SVM on TAL$\_$ASR.}
%   \label{tab:tal_asr}
%   \centering
%   \begin{tabular}{ccc}
%     \toprule
%     \multicolumn{1}{c}{\textbf{Model}} & \multicolumn{1}{c}{\textbf{Accuracy}}&
%     \multicolumn{1}{c}{\textbf{F1-score}}\\
%     \midrule
% BERT-RCNN  &  {\bf 0.9272} & \textbf{0.9270} \\
% n-gram-SVM baseline &  0.7468 & 0.7460 \\
%     \bottomrule
%   \end{tabular}
% \end{table}

%  \vspace{-0.4cm}
% \begin{table}[t!]
%   \caption{Comparision results on TAL$\_$ASR (code-switching short) dataset.}
%   \label{tab:tal_asr}
%   \centering
%   \begin{tabular}{cccc}
%     \toprule
%     \multicolumn{1}{c}{\textbf{Model}} &
%     \multicolumn{1}{c}{\textbf{EER}}&
%     \multicolumn{1}{c}{\textbf{Accuracy}}&
%     \multicolumn{1}{c}{\textbf{F1-score}}\\
%     \midrule
% BERT-RCNN &TODO &  {\bf 0.9461} & \textbf{0.9456} \\
% n-gram-SVM baseline&TODO &  0.7468 & 0.7460 \\
%     \bottomrule
%   \end{tabular}
% \end{table}

\section{Conclusion}
We proposed a BERT-powered language identification system that can extract better phonological and syntactic features beneficial to language identification. We achieved this by deploying a pipeline that includes a phone recognizer, a customized BERT with phone embedding input, and an RCNN classifier. The results show that the proposed BERT-LID system significantly improves language identification performance, especially on the short-segment and code-switching tasks. Further work can consider investigating more optimized network structures and loss functions.
\section{Acknowledgement}

% % The ISCA Board would like to thank the organizing committees of the past INTERSPEECH conferences for their help and for kindly providing the template files. \\
% % Note to authors: Authors should not use logos in the acknowledgement section; rather authors should acknowledge corporations by naming them only.
This work was supported by the National Key R\&D Program of
China under Grant No. 2020AAA0104500, and the National Natural Science Foundation of China under Grant No. U1836219 and No. 62276153.

\bibliographystyle{IEEEtran}

\bibliography{template}

\begin{thebibliography}{10}
\providecommand{\url}[1]{#1}
\csname url@samestyle\endcsname
\providecommand{\newblock}{\relax}
\providecommand{\bibinfo}[2]{#2}
\providecommand{\BIBentrySTDinterwordspacing}{\spaceskip=0pt\relax}
\providecommand{\BIBentryALTinterwordstretchfactor}{4}
\providecommand{\BIBentryALTinterwordspacing}{\spaceskip=\fontdimen2\font plus
\BIBentryALTinterwordstretchfactor\fontdimen3\font minus
  \fontdimen4\font\relax}
\providecommand{\BIBforeignlanguage}[2]{{%
\expandafter\ifx\csname l@#1\endcsname\relax
\typeout{** WARNING: IEEEtran.bst: No hyphenation pattern has been}%
\typeout{** loaded for the language `#1'. Using the pattern for}%
\typeout{** the default language instead.}%
\else
\language=\csname l@#1\endcsname
\fi
#2}}
\providecommand{\BIBdecl}{\relax}
\BIBdecl

\bibitem{albadr2019spoken}
M.~A.~A. Albadr, S.~Tiun, M.~Ayob, and F.~T. AL-Dhief, ``Spoken language
  identification based on optimised genetic algorithm--extreme learning machine
  approach,'' \emph{International Journal of Speech Technology}, vol.~22,
  no.~3, pp. 711--727, 2019.

\bibitem{snyder2018spoken}
D.~Snyder, D.~Garcia-Romero, A.~McCree, G.~Sell, D.~Povey, and S.~Khudanpur,
  ``Spoken language recognition using x-vectors.'' in \emph{Odyssey}, 2018, pp.
  105--111.

\bibitem{waibel2008spoken}
A.~Waibel and C.~Fugen, ``Spoken language translation,'' \emph{IEEE Signal
  Processing Magazine}, vol.~25, no.~3, pp. 70--79, 2008.

\bibitem{di2019adapting}
M.~A. Di~Gangi, M.~Negri, and M.~Turchi, ``Adapting transformer to end-to-end
  spoken language translation,'' in \emph{Proc. INTERSPEECH 2019}.\hskip 1em
  plus 0.5em minus 0.4em\relax ISCA, 2019, pp. 1133--1137.

\bibitem{soltau2016neural}
H.~Soltau, H.~Liao, and H.~Sak, ``Neural speech recognizer: Acoustic-to-word
  lstm model for large vocabulary speech recognition,'' \emph{arXiv preprint
  arXiv:1610.09975}, 2016.

\bibitem{battenberg2017exploring}
E.~Battenberg, J.~Chen, R.~Child, A.~Coates, Y.~G.~Y. Li, H.~Liu, S.~Satheesh,
  A.~Sriram, and Z.~Zhu, ``Exploring neural transducers for end-to-end speech
  recognition,'' in \emph{Proc. ASRU 2017}.\hskip 1em plus 0.5em minus
  0.4em\relax IEEE, 2017, pp. 206--213.

\bibitem{lyu2015mandarin}
D.-C. Lyu, T.-P. Tan, E.-S. Chng, and H.~Li, ``Mandarin--english code-switching
  speech corpus in south-east asia: Seame,'' \emph{Language Resources and
  Evaluation}, vol.~49, no.~3, pp. 581--600, 2015.

\bibitem{nilep2006code}
C.~Nilep, ``“code switching” in sociocultural linguistics,'' \emph{Colorado
  Research in Linguistics}, vol.~19, pp. 1--22, 2006.

\bibitem{chan2006automatic}
J.~Y. Chan, P.~Ching, T.~Lee, and H.~Cao, ``Automatic speech recognition of
  {Cantonese-English} code-mixing utterances,'' in \emph{Proc. ICSLP}, 2006.

\bibitem{li2019towards}
K.~Li, J.~Li, G.~Ye, R.~Zhao, and Y.~Gong, ``Towards code-switching {ASR} for
  end-to-end {CTC} models,'' in \emph{Proc. ICASSP 2019}.\hskip 1em plus 0.5em
  minus 0.4em\relax IEEE, 2019, pp. 6076--6080.

\bibitem{duroselle2021modeling}
R.~Duroselle, M.~Sahidullah, D.~Jouvet, and I.~Illina, ``Modeling and training
  strategies for language recognition systems,'' in \emph{Proc. INTERSPEECH
  2021}, 2021.

\bibitem{liu2021unified}
D.~Liu, J.~Xu, P.~Zhang, and Y.~Yan, ``A unified system for multilingual speech
  recognition and language identification,'' \emph{Speech Communication}, vol.
  127, pp. 17--28, 2021.

\bibitem{yu2003chinese}
S.~Yu, S.~Hu, S.~Zhang, and B.~Xu, ``Chinese-english bilingual speech
  recognition,'' in \emph{Proc. International Conference on Natural Language
  Processing and Knowledge Engineering}.\hskip 1em plus 0.5em minus 0.4em\relax
  IEEE, 2003, pp. 603--609.

\bibitem{lyu2008language}
D.-C. Lyu and R.-Y. Lyu, ``Language identification on code-switching utterances
  using multiple cues,'' in \emph{Proc. INTERSPEECH}, 2008.

\bibitem{snyder2018x}
D.~Snyder, D.~Garcia-Romero, G.~Sell, D.~Povey, and S.~Khudanpur, ``X-vectors:
  Robust dnn embeddings for speaker recognition,'' in \emph{Proc. ICASSP
  2018}.\hskip 1em plus 0.5em minus 0.4em\relax IEEE, 2018, pp. 5329--5333.

\bibitem{wang2006multi}
L.~Wang, E.~Ambikairajah, and E.~H. Choi, ``Multi-lingual phoneme recognition
  and language identification using phonotactic information,'' in \emph{Proc.
  ICPR'06}, vol.~4.\hskip 1em plus 0.5em minus 0.4em\relax IEEE, 2006, pp.
  245--248.

\bibitem{song2015deep}
Y.~Song, X.~Hong, B.~Jiang, R.~Cui, I.~McLoughlin, and L.-R. Dai, ``Deep
  bottleneck network based i-vector representation for language
  identification,'' in \emph{Proc. INTERSPEECH}, 2015.

\bibitem{zhan2021self}
Q.~Zhan, X.~Xie, C.~Hu, and H.~Cheng, ``A self-supervised model for language
  identification integrating phonological knowledge,'' \emph{Electronics},
  vol.~10, no.~18, p. 2259, 2021.

\bibitem{dehak2010front}
N.~Dehak, P.~J. Kenny, R.~Dehak, P.~Dumouchel, and P.~Ouellet, ``Front-end
  factor analysis for speaker verification,'' \emph{IEEE Transactions on Audio,
  Speech, and Language Processing}, vol.~19, no.~4, pp. 788--798, 2010.

\bibitem{li2006vector}
H.~Li, B.~Ma, and C.-H. Lee, ``A vector space modeling approach to spoken
  language identification,'' \emph{IEEE Transactions on Audio, Speech, and
  Language Processing}, vol.~15, no.~1, pp. 271--284, 2006.

\bibitem{zhang2014spoken}
W.-Q. Zhang, W.-W. Liu, Z.-Y. Li, Y.-Z. Shi, and J.~Liu, ``Spoken language
  recognition based on gap-weighted subsequence kernels,'' \emph{Speech
  Communication}, vol.~60, pp. 1--12, 2014.

\bibitem{lee2020subspace}
H.-S. Lee, Y.~Tsao, S.-K. Jeng, and H.-M. Wang, ``Subspace-based representation
  and learning for phonotactic spoken language recognition,'' \emph{IEEE/ACM
  Transactions on Audio, Speech, and Language Processing}, vol.~28, pp.
  3065--3079, 2020.

\bibitem{salamea2016use}
C.~R. Salamea~Palacios, L.~F. D'Haro~Enr{\'\i}quez, R.~d. Cordoba~Herralde, and
  R.~San Segundo~Hern{\'a}ndez, ``On the use of phone-gram units in recurrent
  neural networks for language identification,'' in \emph{Proc. Odyssey}, 2016,
  pp. 117--123.

\bibitem{salamea2018language}
C.~R. Salamea, L.~D'Haro, and R.~Cordoba, ``Language recognition using neural
  phone embeddings and rnnlms,'' \emph{IEEE Latin America Transactions},
  vol.~16, no.~7, pp. 2033--2039, 2018.

\bibitem{romero2021exploring}
D.~Romero, L.~F. D’Haro, and C.~Salamea, ``Exploring transformer-based
  language recognition using phonotactic information,'' \emph{Proc.
  IberSPEECH}, pp. 250--254, 2021.

\bibitem{devlin2018bert}
J.~Devlin, M.-W. Chang, K.~Lee, and K.~Toutanova, ``Bert: Pre-training of deep
  bidirectional transformers for language understanding,'' \emph{arXiv preprint
  arXiv:1810.04805}, 2018.

\bibitem{lai2015recurrent}
S.~Lai, L.~Xu, K.~Liu, and J.~Zhao, ``Recurrent convolutional neural networks
  for text classification,'' in \emph{Proc. Twenty-ninth AAAI conference on
  artificial intelligence}, 2015.

\bibitem{fer2017multilingually}
R.~Fer, P.~Mat{\v{e}}jka, F.~Gr{\'e}zl, O.~Plchot, K.~Vesel{\`y}, and J.~H.
  {\v{C}}ernock{\`y}, ``Multilingually trained bottleneck features in spoken
  language recognition,'' \emph{Computer Speech \& Language}, vol.~46, pp.
  252--267, 2017.

\bibitem{hazen2009query}
T.~J. Hazen, W.~Shen, and C.~White, ``Query-by-example spoken term detection
  using phonetic posteriorgram templates,'' in \emph{Proc. ASRU}.\hskip 1em
  plus 0.5em minus 0.4em\relax IEEE, 2009, pp. 421--426.

\bibitem{sivaram2011multilayer}
G.~S. Sivaram and H.~Hermansky, ``Multilayer perceptron with sparse hidden
  outputs for phoneme recognition,'' in \emph{Proc. ICASSP 2011}.\hskip 1em
  plus 0.5em minus 0.4em\relax IEEE, 2011, pp. 5336--5339.

\bibitem{krizhevsky2012imagenet}
A.~Krizhevsky, I.~Sutskever, and G.~E. Hinton, ``Imagenet classification with
  deep convolutional neural networks,'' \emph{Advances in Neural Information
  Processing Systems}, vol.~25, 2012.

\bibitem{hochreiter1997long}
S.~Hochreiter and J.~Schmidhuber, ``Long short-term memory,'' \emph{Neural
  computation}, vol.~9, no.~8, pp. 1735--1780, 1997.

\bibitem{johnson2017deep}
R.~Johnson and T.~Zhang, ``Deep pyramid convolutional neural networks for text
  categorization,'' in \emph{Proc. 55th Annual Meeting of the Association for
  Computational Linguistics (Volume 1: Long Papers)}, 2017, pp. 562--570.

\bibitem{li2020ap20}
Z.~Li, M.~Zhao, Q.~Hong, L.~Li, Z.~Tang, D.~Wang, L.~Song, and C.~Yang,
  ``Ap20-olr challenge: Three tasks and their baselines,'' in \emph{Proc.
  APSIPA ASC 2020}.\hskip 1em plus 0.5em minus 0.4em\relax IEEE, 2020, pp.
  550--555.

\bibitem{wang2015thchs}
D.~Wang and X.~Zhang, ``Thchs-30: A free chinese speech corpus,'' \emph{arXiv
  preprint arXiv:1512.01882}, 2015.

\bibitem{garofolo1993darpa}
J.~S. Garofolo, L.~F. Lamel, W.~M. Fisher, J.~G. Fiscus, and D.~S. Pallett,
  ``{DARPA TIMIT} acoustic-phonetic continous speech corpus cd-rom. nist speech
  disc 1-1.1,'' \emph{NASA STI/Recon Technical Report N}, vol.~93, p. 27403,
  1993.

\end{thebibliography}

\end{document}